\documentclass[conference,dvipdfmx]{IEEEtran}
\IEEEoverridecommandlockouts
\usepackage{cite}
\usepackage{amsmath,amssymb,amsfonts}
\usepackage{algorithmic}
\usepackage[dvipdfmx]{graphicx}
\usepackage[dvipdfmx]{color}
\usepackage{textcomp}
\usepackage{xcolor}

\usepackage{listings}

\def \TRAINTIME {74.7 seconds}
\def \TRAINTHROUGHPUT {1.73 million}
\def \ACCURACY {75.08\%}

\usepackage{url}
\def\BibTeX{{\rm B\kern-.05em{\sc i\kern-.025em b}\kern-.08em
    T\kern-.1667em\lower.7ex\hbox{E}\kern-.125emX}}
\begin{document}

\title{Yet Another Accelerated SGD: ResNet-50 Training on ImageNet in {\TRAINTIME}}

\author{\IEEEauthorblockN{Masafumi Yamazaki, Akihiko Kasagi, Akihiro Tabuchi, Takumi Honda, Masahiro Miwa,\\
Naoto Fukumoto, Tsuguchika Tabaru, Atsushi Ike, Kohta Nakashima}
\textit{Fujitsu Laboratories Ltd.}\\
\{m.yamazaki, kasagi.akihiko, tabuchi.akihiro, honda.takumi, masahiro.miwa,\\
fukumoto.naoto, tabaru, ike, nakashima.kouta\}@fujitsu.com\\
}

\maketitle

\begin{abstract}
There has been a strong demand for algorithms that can execute machine learning as faster as possible
and the speed of deep learning has accelerated by 30 times only in the past two years. 
Distributed deep learning using the large mini-batch is a key technology to address the demand and is a great
challenge as it is difficult to achieve high scalability on large clusters without compromising accuracy.
In this paper, we introduce optimization methods which we applied to this challenge. 
We achieved the training time of {\TRAINTIME} using 2,048 GPUs on ABCI cluster applying these methods.
The training throughput is over {\TRAINTHROUGHPUT} images/sec and the top-1 validation accuracy is {\ACCURACY}.
\end{abstract}

\section{Introduction}

Deep neural network (DNN) models trained on large datasets are delivering impressive results in various fields,
such as object detection, language translation and so on. However, the computation cost of DNN training
becomes larger since the sizes of DNN models and datasets increase.
Distributed deep learning with data parallelism is known to be an effective approach to accelerate the training
on clusters. In this approach, all processes launched on the cluster have the same DNN model and weights.
Each process trains the model with different mini-batches but the weight gradients from all processes are combined to update all the weights. 
This communication overhead becomes a significant problem for large clusters.
In order to reduce the overhead on large clusters, 
we increase mini-batch size of DNN and compute DNN trainings in parallel.
However, the training with large mini-batch generally results in the worse validation accuracy of DNN models.
Thus, we used several techniques to increase mini-batch size, which denotes the number of input images computed in an iteration, without compromising validation accuracy.


We performed our experimental result, using 2,048 GPUs of AI Bridging Cloud Infrastructure (ABCI) cluster
and self-optimized MXNet deep learning framework.
We achieved {\ACCURACY} validation accuracy of ResNet-50 on ImageNet using 81,920 mini-batch size in {\TRAINTIME}.

\begin{table*}
\caption{Training time and Top-1 validation accuracy with ResNet-50 on ImageNet}
\begin{center}
\begin{tabular}{lccccc}
\hline
          & Batch & Processor &    DL     & Time & Accuracy\\
          & Size   &                 &  Library  &          &               \\
\hline
He et al.~\cite{He2016} & 256    & Tesla P100 $\times$ 8   & Caffe & 29 hours & 75.3 \% \\     
Goyal et al.~\cite{Goyal2017} &  8,192   & Tesla P100 $\times$ 256   & Caffe2 & 1 hour & 76.3 \% \\     
Smith et al.~\cite{Smith2017} & 8,192 $\rightarrow $ 16,384    &  full TPU Pod   & TensorFlow & 30 mins & 76.1 \% \\     
Akiba et al.~\cite{Akiba2017} & 32,768    & Tesla P100 $\times$ 1,024   & Chainer & 15 mins & 74.9 \% \\     
Jia et al.~\cite{Jia2018} & 65,536    & Tesla P40 $\times$ 2,048   & TensorFlow & 6.6 mins & 75.8 \% \\     
Ying et al.~\cite{Ying2018} & 65,536   &  TPU v3 $\times$ 1,024   & TensorFlow & 1.8 mins & 75.2 \% \\     
Mikami et al.~\cite{Mikami2019} & 55,296    & Tesla V100 $\times$ 3,456   & NNL & 2.0 mins & 75.29 \% \\     
{\bf This work} & {\bf 81,920}  & {\bf Tesla V100 $\times$ 2,048}   & {\bf MXNet} & {\bf 1.2 mins} & {\bf \ACCURACY}\\     
\hline
\end{tabular}
\label{tab1}
\end{center}
\end{table*}

\section{Related works}
This section introduces the related works about the large mini-batch challenges.
Alex et al.~\cite{Alex2012} achieved high accuracy for the image recognition in ILSVRC.
This paper shows that convolutional layers are effective for 2D image deep neural network.
Other models which appeared after~\cite{Alex2012} commonly use convolutional layers for 2D and 3D image data.
Ioffe et al.~\cite{Ioffe2015} introduced the batch normalization technique, in which the feature values in hidden layers are normalized to avoid vanishing gradients.
In addition, this technique enables training of models with a large number of layers, such as ResNet.

Generally, the mini-batch size should be large for distributed deep learning on large clusters.
Goyal et al.~\cite{Goyal2017} proposed the warm-up technique to keep the validation accuracy with 8,192 mini-batch size.
Google~\cite{Smith2017} and Sony~\cite{Mikami2019} used the variable mini-batch size which becomes larger and achieved highly parallel processing.

Hence the difference between the weight gradient norm and the weight norm of each layer causes the unstable of the training,
LARS of \cite{You2017} normalizes the difference of each layer 
and the DNN can train with 32,768 without the loss of validation accuracy.

Akiba et al.~\cite{Akiba2017} achieved ResNet-50 training in 15 minutes using 1,024 GPUs. Jia et al.~\cite{Jia2018} also achieved ResNet-50 training in 6.6 minutes using 2,048 GPUs.
Ying et al.~\cite{Ying2018} achieved 1.05 million images/sec by using 1,024 TPU v3 processors. 
The training times of ResNet-50 with 32,768 and 65,536 mini-batch sizes are 2.2 and 1.8 minutes.
These results are summarized in the table~\ref{tab1}.

\section{Our approach}

In this section, we introduce our techniques applied to improve both accuracy and training throughput. 

\subsection{Accuracy Improvement}
We used Stochastic Gradient Descent (SGD) that is commonly used for deep learning optimizer.
When training on large mini-batch, the number of SGD updates decreases as mini-batch size increases, so improving final validation accuracy on large mini-batch is a big challenge, and we adopted the following techniques.

\subsubsection{Learning Rate Control}
We need to use high learning rate to accelerate training due to the small number of updates.
However, high learning rate makes training of models unstable in early stages.
Thus, we stabilize SGD by using the warm-up~\cite{Goyal2017} which raises leaning rate gradually.
Moreover, the same learning rate of all layer is too high for some layers, we also stabilize training by using Layer-wise Adaptive Rate Scaling (LARS)~\cite{You2017}
that adjusts the learning rate of each layer according to the norms weight and gradient.

For convergence of weight, we try many decay patterns of learning rate, such as step, polynomial, linear, and so on.
We used optimized decay patterns based on many trials.

\subsubsection{Other techniques}
It is reported that the label smoothing~\cite{Christian2015} improves accuracy with 32,768 mini-batches~\cite{Mikami2019}.
We also adopt this method and confirmed accuracy improvement on 81,920 mini-batch.

The moving averages of mean and variance of batch normalization layers are computed on each process independently, whereas weights are synchronized.
These values become inaccurate on large mini-batch training~\cite{Akiba2017}; therefore, we tuned some hyper-parameters to optimize the moving averages.

\subsection{Framework optimizations}
We employed MXNet, an open source deep learning framework written in C++ and CUDA C languages, with many language bindings.
MXNet has both flexibility and scalability which enable to train models efficiently on clusters.
However, a part of processing which occupy only a small fraction of total time in small or medium-sized cluster environment may become bottleneck in mass cluster environment. 
We analyzed CPU and GPU performances using several profilers and found out the bottlenecks.
We optimized the bottlenecks to improve the training throughput as following.

\subsubsection{Parallel DNN model initialization}
In data parallel distributed deep learning, 
all layers must be initialized so that these weights have the same values among all processes.
Generally, the root process initializes all weights of the model.
After that, the process broadcasts these weights to all processes.
The broadcast time is increasing in accordance with the number of processes, 
and this broadcast operation cost is not ignored when distributed deep learning by thousands of processes.
Therefore, we employ other initialization approach that every process has the same seed and initializes weights in parallel. 
This approach can synchronize initial weights without the broadcast operation.

\subsubsection{Batched norm computations on the GPU}
The norm computation for each layer is necessary to update the weights with LARS. 
Most of layers of ResNet-50 do not have enough number of weights compared with the number of cores on the GPU.
If we compute the weight norm of each layer on GPU, the number of threads is not enough to occupy all CUDA cores.
Therefore, we implemented a special GPU kernel for batched norm computations into MXNet. 
This GPU kernel can launch enough number of threads, and the norm of layers can be calculated in parallel.

\subsection{Communication Optimizations}

Distributed parallel deep learning requires allreduce communications to  
exchange gradients of each layer between all processes.
Allreduce communication overhead is not negligible in large cluster environment 
because communication time becomes longer while calculation time becomes shorter 
due to the small batch size per GPU.
To overcome these issues, we adopted the following two optimizations.

\subsubsection{Adjusting data size of communication}

Deep learning models are composed of many layers and the data size of gradients varies from layers to layers.
Allreduce operation per each layer leads to large overhead due to frequent callings to communication operation 
and it becomes worse if the data size of gradient is small because network bandwidth cannot be used effectively.
Therefore, it is important to enlarge the data size of allreduce.
We gathered gradients of layers and adjusted the data size of allreduce to several megabytes.

\subsubsection{Optimal scheduling of communications}

We start to operate allreduce operation for a part of layers without waiting all layers to be finished.
This enables allreduce operation to be overlapped with backward processing.
The timing to start the allreduce operation is when the data size of gradients becomes larger than a threshold.
Each process needs to keep pace with the other processes because each processes can only send the gradients of same layers. 
It is possible to find completed layers in common using allgather operation, however this results in additional overhead.
To remove this overhead, we statistically group layers into several groups beforehand. 
Allreduce operation is scheduled as soon as each process finishes backward processing of all layers in a group.

\section{Environment and experimental result}
\begin{figure}[htbp]
\centering
\includegraphics[keepaspectratio=true, scale=0.50]{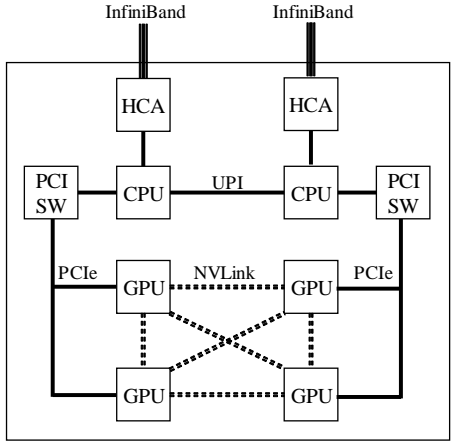}
\caption{The schematic draw of one compute node in ABCI cluster.  It consists of two CPUs, four GPUs and two HCAs connected to corresponding CPUs.}
\label{ABCI_node}
\end{figure}

We used ABCI cluster to evaluate the performance of our optimized framework based on MXNet.
Each node of ABCI cluster consists of two CPUs of Xeon Gold 6148 and four GPUs of NVIDIA Tesla V100 SXM2.
In addition, GPUs on a node are connected by NVLink and nodes also have two InfiniBand Network Interface Cards.
Fig.~\ref{ABCI_node} shows the architecture of a node of ABCI cluster.

We used mixed precision method, where we compute and communicate using half precision floating point numbers and update own weights using single precision floating point numbers.
We used our original optimizer which enables fine control of learning rate.
In addition to stabilizing the training accuracy, we also used warmup~\cite{Goyal2017} and LARS~\cite{You2017} techniques.

Our measurement of ResNet-50 training is according to the rule of MLPerf v0.5.0.
This means that we measure the elapsed time from the message of ``run\_start'' to ``run\_final''
which includes both initialization and memory allocation time.

Appendix~\ref{result_log} is from the actual output logs of our experiments.
As shown in this results, our optimized DNN framework achieved
completing the ResNet-50 training on ImageNet in {\TRAINTIME} with {\ACCURACY} validation accuracy.

\begin{figure}[htbp]
\centering
\includegraphics[keepaspectratio=true, scale=0.75]{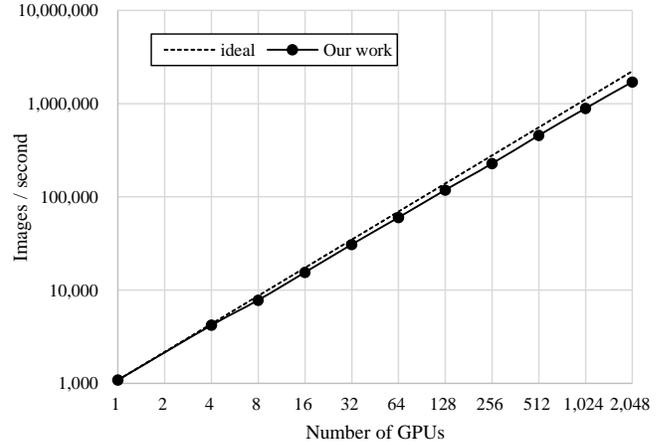}
\caption{The scalability of our optimized framework shown by the solid line.  The dashed line shows the ideal curve.}
\label{scalability}
\end{figure}

We also measured the scalability of ResNet-50.
Fig.~\ref{scalability} shows the computational throughput according to the number of GPUs.
In Fig.~\ref{scalability}, the dotted line denotes the ideal throughput of images-per-second, and the solid line denotes our result which shows the scalability of our framework is quite good until 2,048 GPUs.
The throughput using 2,048 GPUs is 1.7 million images-per-second and the scalability is 77.0~\%.

\begin{figure}[htbp]
\centering
\includegraphics[keepaspectratio=true, scale=0.75]{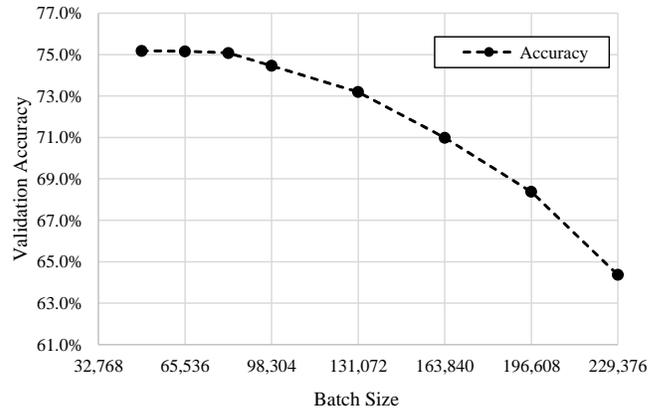}
\caption{The top-1 validation accuracies in 49,152 or larger mini-batch training}
\label{accuracy}
\end{figure}

Fig.~\ref{accuracy} shows the result of top-1 validation accuracies in 81,920 or larger mini-batch training.
In this Fig.~\ref{accuracy}, the validation accuracies over 81,920 mini-batches is lower than 74.9~\%, which cannot meet to MLPerf regulation.
Hence the number of images in one epochs of ImageNet dataset is 1,280,000 images,
the number of updates in an epoch is only 16 if we use 81,920 mini-batches, where the number of total update count is 1,440.
This number is too small for SGD solvers to train the DNN weights.
Thus, using large mini-batches is a big challenge and we tried to use as large mini-batch as possible.
As shown in the Table~\ref{tab1}, 81,920 mini-batch size is so large comparing to other works and we reached to over 75~\% of the validation accuracy.

\begin{figure}[htbp]
\centering
\includegraphics[keepaspectratio=true, scale=0.75]{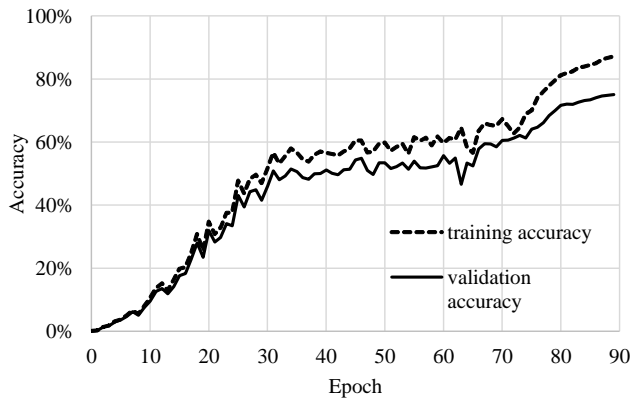}
\caption{The comparison between training accuracy and validation accuracy.}
\label{train_vs_test}
\end{figure}

Fig.~\ref{train_vs_test} is the comparison between training and validation accuracy.
This figure shows that our results of validation accuracy is not overfitting by using batch normalization and label smoothing techniques.



\section{Conclusion}
We developed the novel techniques to use large mini-batch on large scale GPU clusters without loss of validation accuracy.
We applied the technique to our deep learning framework based on MXNet.
The result of our DNN training achieve {\ACCURACY} validation accuracy of ResNet-50 using 81,920 mini-batch size in {\TRAINTIME}.

\section{Future work}

The training of large DNN, such as ResNet-50, which once required a large amount of time, can now execute
only in a minute or so.
Thanks to this, it has become possible to perform various trials in a short time and we could have
obtained the several techniques to gain the validation accuracy.
As a next step, we will try to formulate the techniques and apply the contributions to the general DNN trainings.

\section{Acknowledgements}
We'd like to thank N. Hamada, K. Uemura, T. Kato, K. Matsumoto, Y. Kanazawa, H. Ohtsuji, K. Yoda, Y. Sakai, K. Tsugane, K. Shirahata and Y. Tomita for the helpful discussions
on theoretical background.
We'd also like to thank to National Institute of Advanced Industrial Science and Technology (AIST) and their support team for the stable use of ABCI in a large cluster of 512 nodes.

\bibliography{fj_mxnet.bib}
\bibliographystyle{unsrt}

\onecolumn
\appendix

\begin{lstlisting}[basicstyle=\ttfamily\footnotesize, frame=single, caption=The log file of ResNet50 training with 512 nodes and 2048 GPUs, label=result_log]
\caption{}
:::MLPv0.5.0 resnet 1553154085.031997204 (/fs3/home/aca10034mq/mxnet/JobScripts/image_classification/mlperf_log_utils.py:69) eval_offset: 1
:::MLPv0.5.0 resnet 1553154085.032542229 (/fs3/home/aca10034mq/mxnet/JobScripts/image_classification/mlperf_log_utils.py:69) run_start
:::MLPv0.5.0 resnet 1553154085.032880306 (/fs3/home/aca10034mq/mxnet/JobScripts/image_classification/mlperf_log_utils.py:69) run_set_random_seed: 100000
:::MLPv0.5.0 resnet 1553154085.040173054 (/fs3/home/aca10034mq/mxnet/JobScripts/image_classification/mlperf_log_utils.py:69) model_hp_initial_shape: [4, 224, 224]
:::MLPv0.5.0 resnet 1553154085.041487932 (/fs3/home/aca10034mq/mxnet/JobScripts/image_classification/mlperf_log_utils.py:69) model_hp_conv2d_fixed_padding: "[4, 224, 224] -> (64, 112.0, 112.0)"
:::MLPv0.5.0 resnet 1553154085.042191505 (/fs3/home/aca10034mq/mxnet/JobScripts/image_classification/mlperf_log_utils.py:69) model_hp_conv2d_fixed_padding: {"stride": 2, "filters": 64, "initializer": "truncated_normal", "use_bias": false}
:::MLPv0.5.0 resnet 1553154085.043102980 (/fs3/home/aca10034mq/mxnet/JobScripts/image_classification/mlperf_log_utils.py:69) model_hp_batch_norm: {"shape": [64, 112.0, 112.0], "momentum": 0.9, "epsilon": 1e-05, "center": true, "scale": true, "training": true}
:::MLPv0.5.0 resnet 1553154085.044021845 (/fs3/home/aca10034mq/mxnet/JobScripts/image_classification/mlperf_log_utils.py:69) model_hp_initial_max_pool: "(64, 112.0, 112.0) -> (64, 56.0, 56.0)"
:::MLPv0.5.0 resnet 1553154085.044733763 (/fs3/home/aca10034mq/mxnet/JobScripts/image_classification/mlperf_log_utils.py:69) model_hp_begin_block: {"block_type": "bottleneck_block"}
...
:::MLPv0.5.0 resnet 1553154091.058884621 (/fs3/home/aca10034mq/mxnet/JobScripts/image_classification/mlperf_log_utils.py:69) train_loop
:::MLPv0.5.0 resnet 1553154091.185450077 (/fs3/home/aca10034mq/mxnet/JobScripts/image_classification/mlperf_log_utils.py:69) train_epoch: 0
:::MLPv0.5.0 resnet 1553154092.852500200 (/fs3/home/aca10034mq/mxnet/JobScripts/image_classification/mlperf_log_utils.py:69) train_epoch: 1
:::MLPv0.5.0 resnet 1553154093.741370916 (/fs3/home/aca10034mq/mxnet/JobScripts/image_classification/mlperf_log_utils.py:69) eval_start
:::MLPv0.5.0 resnet 1553154093.815561533 (/fs3/home/aca10034mq/mxnet/JobScripts/image_classification/mlperf_log_utils.py:69) eval_accuracy: {"epoch": 1, "value": 0.00289}
:::MLPv0.5.0 resnet 1553154093.816023827 (/fs3/home/aca10034mq/mxnet/JobScripts/image_classification/mlperf_log_utils.py:69) eval_stop
:::MLPv0.5.0 resnet 1553154093.816426039 (/fs3/home/aca10034mq/mxnet/JobScripts/image_classification/mlperf_log_utils.py:69) train_epoch: 2
:::MLPv0.5.0 resnet 1553154094.580265045 (/fs3/home/aca10034mq/mxnet/JobScripts/image_classification/mlperf_log_utils.py:69) train_epoch: 3
:::MLPv0.5.0 resnet 1553154095.400528193 (/fs3/home/aca10034mq/mxnet/JobScripts/image_classification/mlperf_log_utils.py:69) train_epoch: 4
:::MLPv0.5.0 resnet 1553154096.241724491 (/fs3/home/aca10034mq/mxnet/JobScripts/image_classification/mlperf_log_utils.py:69) train_epoch: 5
:::MLPv0.5.0 resnet 1553154096.987241983 (/fs3/home/aca10034mq/mxnet/JobScripts/image_classification/mlperf_log_utils.py:69) eval_start
:::MLPv0.5.0 resnet 1553154097.044126749 (/fs3/home/aca10034mq/mxnet/JobScripts/image_classification/mlperf_log_utils.py:69) eval_accuracy: {"epoch": 5, "value": 0.03604}
:::MLPv0.5.0 resnet 1553154097.044571877 (/fs3/home/aca10034mq/mxnet/JobScripts/image_classification/mlperf_log_utils.py:69) eval_stop
...
:::MLPv0.5.0 resnet 1553154153.491471767 (/fs3/home/aca10034mq/mxnet/JobScripts/image_classification/mlperf_log_utils.py:69) train_epoch: 82
:::MLPv0.5.0 resnet 1553154154.156885147 (/fs3/home/aca10034mq/mxnet/JobScripts/image_classification/mlperf_log_utils.py:69) train_epoch: 83
:::MLPv0.5.0 resnet 1553154154.903352499 (/fs3/home/aca10034mq/mxnet/JobScripts/image_classification/mlperf_log_utils.py:69) train_epoch: 84
:::MLPv0.5.0 resnet 1553154155.639714956 (/fs3/home/aca10034mq/mxnet/JobScripts/image_classification/mlperf_log_utils.py:69) train_epoch: 85
:::MLPv0.5.0 resnet 1553154156.332974434 (/fs3/home/aca10034mq/mxnet/JobScripts/image_classification/mlperf_log_utils.py:69) eval_start
:::MLPv0.5.0 resnet 1553154156.382446527 (/fs3/home/aca10034mq/mxnet/JobScripts/image_classification/mlperf_log_utils.py:69) eval_accuracy: {"epoch": 85, "value": 0.7343}
:::MLPv0.5.0 resnet 1553154156.382866144 (/fs3/home/aca10034mq/mxnet/JobScripts/image_classification/mlperf_log_utils.py:69) eval_stop
:::MLPv0.5.0 resnet 1553154156.383258343 (/fs3/home/aca10034mq/mxnet/JobScripts/image_classification/mlperf_log_utils.py:69) train_epoch: 86
:::MLPv0.5.0 resnet 1553154157.177564383 (/fs3/home/aca10034mq/mxnet/JobScripts/image_classification/mlperf_log_utils.py:69) train_epoch: 87
:::MLPv0.5.0 resnet 1553154158.042360306 (/fs3/home/aca10034mq/mxnet/JobScripts/image_classification/mlperf_log_utils.py:69) train_epoch: 88
:::MLPv0.5.0 resnet 1553154158.825143576 (/fs3/home/aca10034mq/mxnet/JobScripts/image_classification/mlperf_log_utils.py:69) train_epoch: 89
:::MLPv0.5.0 resnet 1553154159.642317295 (/fs3/home/aca10034mq/mxnet/JobScripts/image_classification/mlperf_log_utils.py:69) eval_start
:::MLPv0.5.0 resnet 1553154159.685859919 (/fs3/home/aca10034mq/mxnet/JobScripts/image_classification/mlperf_log_utils.py:69) eval_accuracy: {"epoch": 89, "value": 0.75082}
:::MLPv0.5.0 resnet 1553154159.686291456 (/fs3/home/aca10034mq/mxnet/JobScripts/image_classification/mlperf_log_utils.py:69) eval_stop
:::MLPv0.5.0 resnet 1553154159.686674595 (/fs3/home/aca10034mq/mxnet/JobScripts/image_classification/mlperf_log_utils.py:69) run_stop
:::MLPv0.5.0 resnet 1553154159.687013626 (/fs3/home/aca10034mq/mxnet/JobScripts/image_classification/mlperf_log_utils.py:69) run_final
 \end{lstlisting}

\end{document}